\documentclass[11pt]{article}

\usepackage[preprint]{acl}

\usepackage{times}
\usepackage{latexsym}

\usepackage[T1]{fontenc}

\usepackage[utf8]{inputenc}

\usepackage{microtype}

\usepackage{inconsolata}

\usepackage{graphicx}

\usepackage{booktabs}       
\usepackage{amsfonts}       
\usepackage{nicefrac}       
\usepackage{xcolor}         
\usepackage{multirow}
\usepackage{amsmath}
\usepackage{bm}
\usepackage{xspace}
\usepackage{colortbl}
\usepackage{utfsym}
\usepackage{amssymb}
\usepackage{pifont}
\usepackage{makecell}
\usepackage{tcolorbox}

\title{GeoAlign: Geometric Feature Realignment for MLLM Spatial Reasoning}

\author{
    \textbf{Zhaochen Liu\textsuperscript{\rm 1}},
    \textbf{Limeng Qiao\textsuperscript{\rm 2}},
    \textbf{Guanglu Wan\textsuperscript{\rm 2}},
    \textbf{Tingting Jiang\textsuperscript{\rm 1,3}\thanks{Corresponding author.}}\\
    \textsuperscript{1}National Engineering Research Center of Visual Technology, National Key Laboratory\\ for Multimedia Information Processing, School of Computer Science, Peking University\\
    \textsuperscript{2}Meituan Inc.
    \textsuperscript{3}National Biomedical Imaging Center, Peking University\\
    \texttt{\{dreamerliu, qiaolm, ttjiang\}@pku.edu.cn}
}

\begin{document}
\maketitle

\begin{abstract}
  Multimodal large language models~(MLLMs) have exhibited remarkable performance in various visual tasks, yet still struggle with spatial reasoning. Recent efforts mitigate this by injecting geometric features from 3D foundation models, but rely on static single-layer extractions. We identify that such an approach induces a task misalignment bias: the geometric features naturally evolve towards 3D pretraining objectives, which may contradict the heterogeneous spatial demands of MLLMs, rendering any single layer fundamentally insufficient. To resolve this, we propose \textbf{GeoAlign}, a novel framework that dynamically aggregates multi-layer geometric features to realign with the actual demands. GeoAlign constructs a hierarchical geometric feature bank and leverages the MLLM's original visual tokens as content-aware queries to perform layer-wise sparse routing, adaptively fetching the suitable geometric features for each patch. Extensive experiments on VSI-Bench, ScanQA, and SQA3D demonstrate that our compact 4B model effectively achieves state-of-the-art performance, even outperforming larger existing MLLMs.
\end{abstract}

\section{Introduction}
\label{sec:intro}

The human visual system inherently perceives the world not merely as a flattened canvas, but as a structured, three-dimensional environment. This innate spatial intelligence allows us to effortlessly judge, comprehend, and interact with the physical world. In contrast, while modern multimodal large language models~(MLLMs) have achieved profound progress in diverse visual tasks, they still struggle when faced with spatial reasoning tasks~\cite{kamathetal2023whatsup, shirietal2024spatialmm, wang2024spatialeval, yang2025thinking}, lacking the intrinsic geometric capabilities.

To enhance the spatial intelligence of MLLMs, early trajectories attempt to introduce explicit 3D representations~\cite{hong20233dllm, zheng2025video3dllm, zhu2025llava3d}, such as point clouds or depth maps. While effective on 3D question-answering benchmarks, the rigid dependency on specialized 3D data limits the scalability on general visual inputs. Recently, a simplified paradigm has emerged: utilizing implicit dense features extracted by feed-forward 3D geometry foundation models~\cite{wang2024dust3r,wang2025vggt}. The extracted features contain rich and compressed geometric content, thereby enabling an efficient framework for spatial reasoning~\cite{zheng2025vgllm, wu2025spatialmllm, yang2025cambrian, fan2026vlm3r}.

\begin{figure}[t]
  \centering
   \includegraphics[width=\columnwidth]{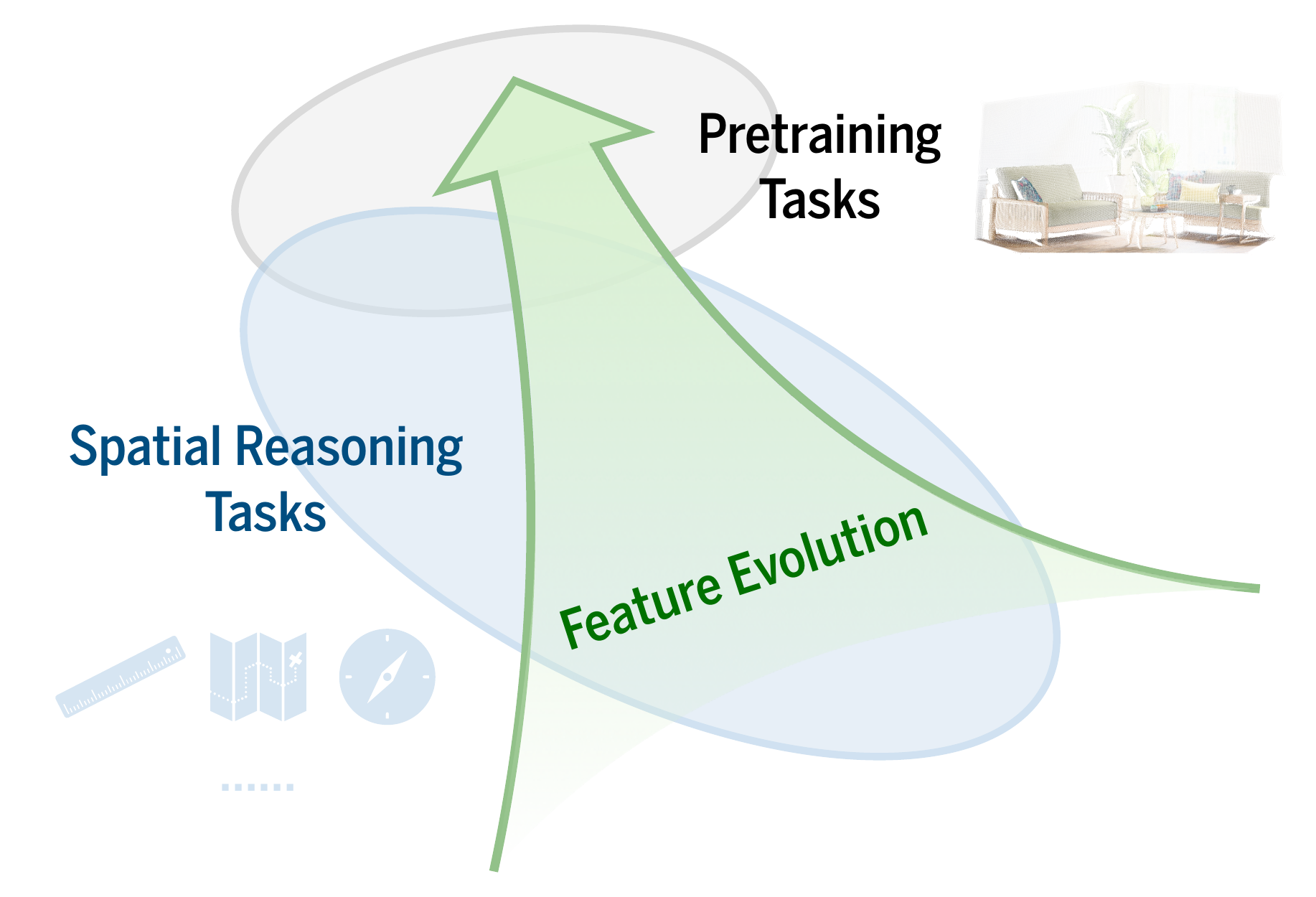}
   \caption{\textbf{Task misalignment bias.} Feature evolution progressively aligns with the pretraining tasks, thus many geometric features valuable for spatial reasoning tasks are distributed in the preceding layers.}
   \label{fig:teaser}
\end{figure}

Despite this progress, current spatial-enhanced MLLMs adopt a static single-layer extraction strategy, fetching features solely from one deep layer of the geometric encoder. However, as features propagate through the geometric encoder, they undergo a gradual transition towards the pretrained tasks~\cite{yosinski2014transferable}, which induces a task misalignment bias. Specifically, feature evolution within the geometric encoder is not consistently beneficial for spatial reasoning tasks. Our empirical study demonstrates that diverse spatial tasks exhibit distinct layer-wise preferences, suggesting that no single layer is sufficient for the complex demands of spatial reasoning.

To resolve this issue and fully harness the potential of the geometric foundation model, we propose \textbf{GeoAlign}, geometric feature realignment for spatial reasoning. GeoAlign abandons the static single-layer paradigm in favor of a dynamic multi-layer aggregation strategy. We first construct a hierarchical feature bank from the geometric encoder, capturing a comprehensive spectrum of geometric content. Subsequently, the original visual tokens are utilized to actively act as content-aware queries. Through a lightweight routing mechanism, they dynamically fetch and aggregate suitable geometric features for each patch. The fused, task-aligned geometric features are then injected into the MLLM's visual stream via a residual pathway.

To evaluate the effectiveness of GeoAlign, we conduct experiments across diverse spatial reasoning and 3D scene understanding benchmarks. Operating at a compact 4B parameter scale, GeoAlign achieves state-of-the-art performance~(71.4) on VSI-Bench~\cite{yang2025thinking}, even significantly surpassing larger MLLMs. On ScanQA~\cite{azuma2022scanqa} and SQA3D~\cite{masqa3d}, GeoAlign also achieves top performance comparable to VLM-3R-7B~\cite{fan2026vlm3r}. Furthermore, our comprehensive ablation studies empirically confirm the performance gains brought by the proposed method and determine the specific architectural configurations.
In summary, our main contributions are threefold:
\begin{itemize}
    \item We identify and empirically validate the task misalignment bias in current spatial-enhanced MLLMs, revealing the limitations of the static single-layer extraction strategy.
    \item We propose GeoAlign, a novel framework that dynamically aggregates multi-layer geometric features to realign with the demands of spatial reasoning tasks.
    \item Extensive experiments are conducted, demonstrating that our compact 4B model effectively yields superior performance, even outperforming larger existing MLLMs.
\end{itemize}

\section{Related Work}
\label{sec:related_work}

\subsection{Multimodal Large Language Models}
The rapid evolution of multimodal large language models has reshaped the general paradigm for visual tasks. By aligning vision encoders with language backbones~\cite{liu2023llava, liu2024llava1.5, li2023blip2}, MLLMs demonstrate remarkable proficiency in visual question-answering and instruction following. However, when confronted with tasks demanding spatial cognition, such as directions, distances, occlusions, or layouts, current MLLMs~\cite{zhang2024llavavideo,li2024llava, bai2025qwen2.5vl, zhu2025internvl3} exhibit inadequate capabilities~\cite{kamathetal2023whatsup, shirietal2024spatialmm, wang2024spatialeval, chen2024spatialvlm, yang2025thinking, wang2025spatial457, yeh2025seeing, wasi2026spatialab}. As a cornerstone towards broader applications, how MLLM captures the underlying geometry of the physical world remains an open issue.

\subsection{3D-Aware MLLMs}
To bridge the gap between 2D semantics and 3D spatial intelligence, researchers attempt to input explicit 3D representations into LLMs. Methods such as 3D-LLM~\cite{hong20233dllm}, LL3DA~\cite{chen2024ll3da}, ChatScene~\cite{zhang2024chatscene}, Video-3D LLM~\cite{zheng2025video3dllm}, and LLaVA-3D~\cite{zhu2025llava3d} incorporate point clouds, voxel grids, or depth maps. While highly effective on 3D question-answering benchmarks~\cite{azuma2022scanqa, masqa3d}, this explicit paradigm introduces additional data from specialized models or sensors, thus limits the scalability and generalization capabilities on common images or videos. 

\subsection{Spatial-Enhanced MLLMs}
To overcome the limitations of explicit 3D inputs, a new trajectory arises to elicit spatial reasoning solely from images or videos. Leveraging implicit dense features extracted by feed-forward 3D geometry foundation models~\cite{wang2024dust3r, wang2025vggt, wang2025pi3}, current works~\cite{zheng2025vgllm, wu2025spatialmllm, yang2025cambrian, fan2026vlm3r} make significant progress in spatial reasoning. 
However, the prevailing injection paradigm exploits static, single-layer geometric features. This paradigm ignores the progressive evolution within the geometric encoder is not entirely consistent with the demands of spatial reasoning. In response, our proposed approach dynamically aggregates multi-layer geometric features, achieving better alignment and superior performance.

\begin{table*}[t]
\centering
\caption{\textbf{Impact of feature layer selection.} We evaluate Qwen2.5-VL finetuned with geometric features from two distinct layers~(Layer-12 and Layer-20) of the VGGT encoder on VSI-Bench. The $\Delta$ row indicates the performance difference~(Layer-20 minus Layer-12). The better results for each task are marked in \textbf{bold}.}
\label{tab:layer_ablation}
\resizebox{\textwidth}{!}{%
\begin{tabular}{l|cccccccc}
\toprule
Feature Source$\ $ & $\ $Route Plan. & Appr. Order & Room Size & Obj. Size & Obj. Count & Rel. Dist. & Abs. Dist. & Rel. Dir. \\
\midrule
Layer-12 & \textbf{47.9} & \textbf{83.5} & \textbf{74.0} & \textbf{74.4} & 69.5 & 66.3 & 55.1 & 83.2 \\
Layer-20    & 43.2 & 79.9 & 72.0 & 73.9 & \textbf{69.9} & \textbf{67.5} & \textbf{59.0} & \textbf{89.0} \\
\midrule
\rowcolor{gray!10}
\textbf{$\Delta$} & -4.7 & -3.6 & -2.0 & -0.5 & +0.4 & +1.2 & +3.9 & +5.8
\\
\bottomrule
\end{tabular}%
}
\end{table*}

\section{Task Misalignment Bias}
Classic representation learning theory establishes that shallow network layers extract generic, universally applicable features, while deeper layers become progressively tailored to the specific objectives of their pre-training tasks~\cite{yosinski2014transferable}. The prevailing paradigm of injecting geometric features into MLLMs relies exclusively on a single predetermined deep layer of the geometric foundation model. We argue that this static strategy suffers from an inherent task misalignment bias: because the objectives optimized during 3D pretraining do not perfectly align with the diverse demands of spatial reasoning, the feature evolution within the geometric foundation model inherently fails to uniformly benefit all types of spatial queries.

To empirically validate this, we conduct an exploratory study. By adopting the paradigm of VG-LLM~\cite{zheng2025vgllm}, we independently add VGGT~\cite{wang2025vggt} features from distinct single layers~(specifically, Layer-12 and Layer-20) to the original visual tokens and finetune the MLLM~(refer to Sec.~\ref{sec:exp_implementation} for detailed implementations). As illustrated in Table~\ref{tab:layer_ablation}, the results on VSI-Bench~\cite{yang2025thinking} reveal a significant divergence in feature preference across the tasks. Certain tasks, such as \emph{route plan}~(47.9\% vs. 43.2\%) and \emph{room size}~(74.0\% vs. 72.0\%), exhibit a clear preference for the earlier Layer-12 features. Conversely, tasks more directly aligning with 3D pretraining objectives, such as \emph{relative distance}~(66.3\% vs. 67.5\%) and \emph{relative direction}~(83.2\% vs. 89.0\%), achieve better performance when utilizing the deeper Layer-20 features.

The observed disparity confirms the layer-wise feature evolution within the geometric foundation model. During pretraining, these models are optimized toward 3D reconstruction objectives, such as dense point map prediction and camera pose estimation. In earlier layers~(e.g., Layer-12), the geometric representations remain relatively generic, thus retaining a broader applicability. Conversely, as features propagate to deeper layers~(e.g., Layer-20), they become specialized to align with the original reconstruction targets. While this specialization benefits spatial tasks that directly demand geometric coordinates, it simultaneously induces an unintended suppression of some generic geometric signals. Consequently, these deep features yield degraded performance compared to their earlier counterparts for certain spatial reasoning tasks.

This fundamental contradiction establishes that no single, static geometric feature layer can universally satisfy the composite demands of spatial reasoning. Recently, the attention residuals mechanism~\cite{chen2026attnres} in large language models~(LLMs) demonstrates that dynamically attending to preceding layer outputs allows models to selectively retrieve and exploit early-layer knowledge, significantly boosting model performance. Inspired by this philosophy of selective layer aggregation, we posit that integrating geometric features into MLLMs must transcend single-layer extraction in favor of a hierarchical fusion mechanism.

\begin{figure*}[t]
  \centering
   \includegraphics[width=0.93\linewidth]{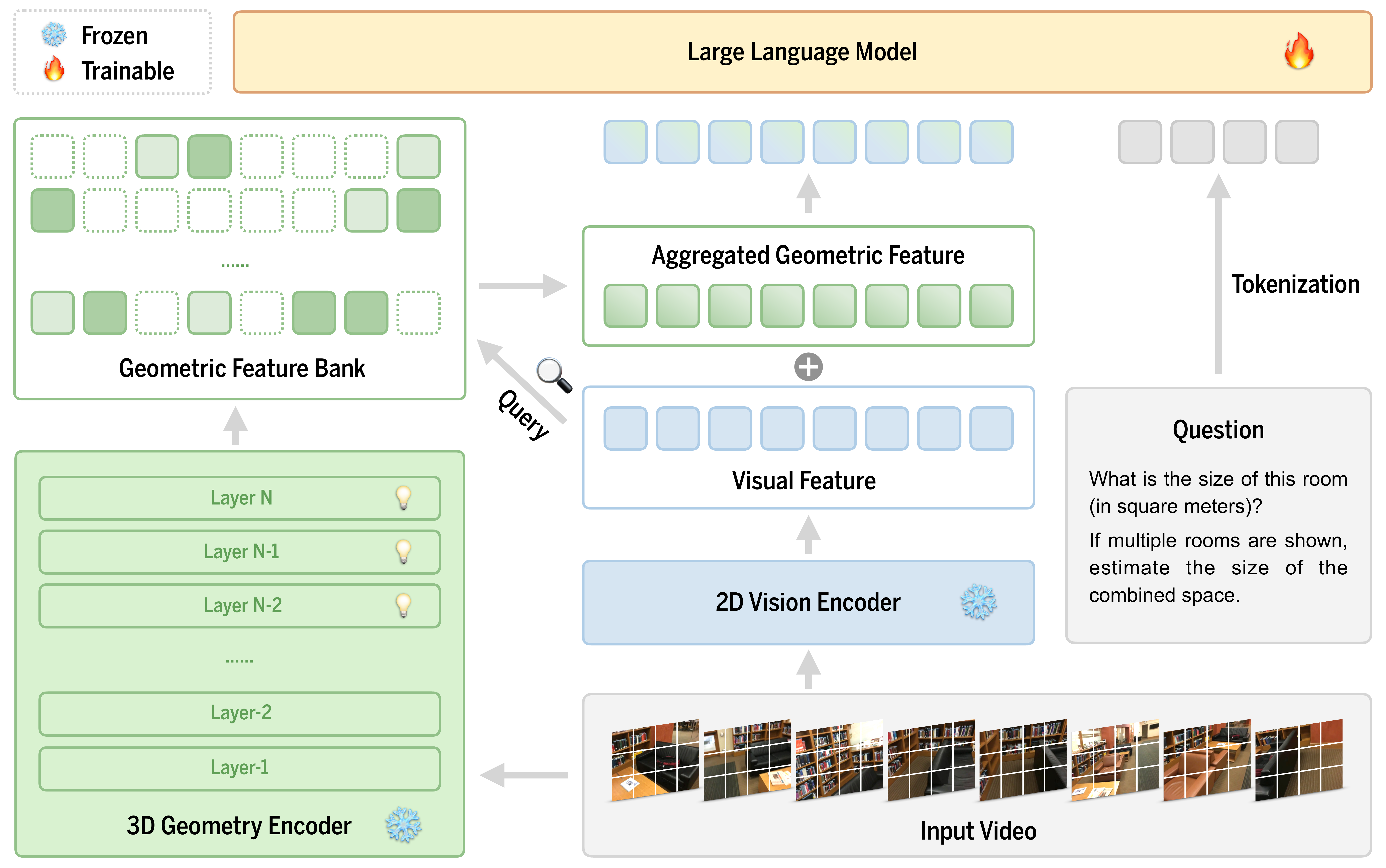}
   \caption{\textbf{Overview of the GeoAlign framework.} We augment the 2D visual features with aggregated geometric features, which are adaptively selected and fused from a hierarchical feature bank built upon the 3D geometry encoder. In this dynamic routing mechanism, the original visual tokens act as content-aware queries, ensuring that the injected geometric features properly align with diverse spatial reasoning demands.}
   \label{fig:pipeline}
\end{figure*}

\section{GeoAlign}
\label{sec:method}

To overcome the inherent task misalignment bias and harness the progressive evolution within the geometric encoder, we propose \textbf{GeoAlign}. As shown in Fig.~\ref{fig:pipeline}, rather than passively accepting a predetermined geometric prior, this mechanism empowers the MLLM to actively query and aggregate suitable geometric features at a per-patch level.

\paragraph{Geometric Feature Bank.}
We first construct a comprehensive geometric feature bank. Given an input visual sequence, we extract multi-layer representations from a continuous subset of $M$ intermediate layers of the geometric foundation model, capturing different stages of the feature evolution. Let $\bm{R}_i \in \mathbb{R}^{L' \times D'}$ denote the raw geometric feature extracted from the $i$-th selected layer, where $L'$ and $D'$ are the native length and dimension of $\bm{R}_i$, respectively. To bridge the semantic modality gap and align with the MLLM's visual feature layout~($L$) and hidden size~($D$), each raw geometric feature undergoes a layer-specific normalization to prevent inter-layer variance pollution, followed by a shared two-layer MLP projection:
\begin{equation}
    \bm{F}_i = f_{\phi} \big( \mathrm{Norm}_i(\bm{R}_i) \big) \in \mathbb{R}^{L \times D},
\end{equation}
where $\mathrm{Norm}_i(\cdot)$ is the LayerNorm assigned to the $i$-th layer, and $f_{\phi}(\cdot)$ is the shared MLP. Subsequently, the geometric feature bank $\mathcal{B}$ is formulated as a stacked tensor of these translated hierarchical representations:
\begin{equation}
    \mathcal{B} = \big[ \bm{F}_1, \bm{F}_2, \dots, \bm{F}_M \big] \in \mathbb{R}^{L \times M \times D}.
\end{equation}

\paragraph{Content-Aware Querying.}
To determine the optimal geometric features required for each patch, we leverage the informative original visual representations from the MLLM's vision encoder as content-aware queries. Let $\bm{Q} \in \mathbb{R}^{L \times D}$ denote the original visual feature sequence. We introduce a routing network $f_{\theta}(\cdot)$, implemented as a lightweight two-layer MLP, to project $\bm{Q}$ into an $M$-dimensional logit space. This MLP explicitly infers the patch-level preference across $M$ candidate layers:
\begin{equation}
    \bm{S} = f_{\theta}(\bm{Q}) \in \mathbb{R}^{L \times M},
\end{equation}
where each element $\bm{S}_{l, i}$ represents the preference score allocated to the $i$-th geometric layer in the feature bank $\mathcal{B}$ by the $l$-th original visual token.

\paragraph{Sparse Aggregation.}
A naive dense aggregation~(e.g., standard $\mathrm{Softmax}$ over all $M$ layers) suffers from training challenges. The accumulated geometric signals from low-weight layers may act as noise that pollutes the semantic manifold and induces structural interference, while the smooth blending reduces the pressure to learn discriminative routing weights. To preserve feature purity and enforce sharp routing decisions, we introduce a hard sparsity constraint. 
For each visual token $l$, we isolate the indices of the $K$ highest-scoring geometric layers~($K \ll M$) using a straightforward selection operator:
\begin{equation}
    \Omega_l = \mathrm{TopK} \big( \{ \bm{S}_{l, i} \}_{i=1}^M \big).
\end{equation}
We then apply a sparsity mask to truncate routing scores of unselected layers, obtaining the masked logits $\hat{\bm{S}}_{l, i}$:
\begin{equation}
    \hat{\bm{S}}_{l, i} = 
    \begin{cases} 
      \bm{S}_{l, i}, & \text{if } i \in \Omega_l \\
      -\infty, & \text{otherwise}
    \end{cases}.
\end{equation}
Subsequently, the masked logits are normalized via the $\mathrm{Softmax}$ function across the candidate layers to yield the sparse routing weights:
\begin{equation}
\bm{\alpha} = \mathrm{Softmax}(\hat{\bm{S}}) \in \mathbb{R}^{L \times M}. 
\end{equation}
The aggregated geometric feature $\hat{\bm{F}}_l$ is synthesized through a weighted fusion of the features across $M$ candidate layers:
\begin{equation}
    \hat{\bm{F}}_l = \sum_{i=1}^{M} \bm{\alpha}_{l, i} \bm{F}_{l, i} \in \mathbb{R}^{D},
\end{equation}
where $\bm{F}_{l, i} \in \mathbb{R}^D$ denotes the geometric feature vector corresponding to the $l$-th visual token from the $i$-th layer in $\mathcal{B}$. The full sequence of the aggregated geometric features is given by 
$\hat{\bm{F}} = [\hat{\bm{F}}_1; \dots; \hat{\bm{F}}_L] \in \mathbb{R}^{L \times D}$. 

\paragraph{Residual Injection.}
Finally, the aggregated geometric feature $\hat{\bm{F}}$ is injected into the visual pathway prior to the LLM backbone. Specifically, we project the geometric feature using a linear transformation $\bm{W}_{out}$, and add it to the original visual feature $\bm{Q}$ via a residual connection:
\begin{equation}
    \hat{\bm{Q}} = \bm{Q} + \bm{W}_{out} \hat{\bm{F}}.
\end{equation}

\begin{table*}[!t]
\centering
\caption{\textbf{Evaluations on VSI-Bench for spatial reasoning.} We compare our proposed GeoAlign model against representative proprietary models, open-sourced models, and spatial-enhanced models. Following the benchmark guidelines, we report the accuracy for multiple-choice tasks, and the mean relative accuracy for numerical tasks. The best performance of each column is marked in \textbf{bold}, and the second best is \underline{underlined}.}
\resizebox{\textwidth}{!}
{
    \begin{tabular}{l|c|cccccccc}
    \toprule
    \multirow{2}{*}{\raisebox{-0.5ex}{\textbf{Models}}}   &
    \multirow{2}{*}{\raisebox{-0.5ex}{$\ $\textbf{Avg.}$\ $}}      &
    \multicolumn{4}{c}{\textbf{Numerical}}        & 
    \multicolumn{4}{c}{\textbf{Multiple-Choice}} \\
    \cmidrule(lr){3-6}\cmidrule(lr){7-10}
    & & $\ $Obj. Cnt. & Abs. Dist. & Obj. Size & Room Size 
    & Rel. Dist. & Rel. Dir. & Route Plan & Appr. Order$\ $ \\
    \midrule
    \rowcolor{gray!10}
    \multicolumn{10}{l}{\textcolor{black}{\textit{Proprietary Models}}} \\
    GPT-4o & 34.0 & 46.2 & 5.3 & 43.8 & 38.2 & 37.0 & 41.3 & 31.5 & 28.5 \\
    Gemini-1.5-Pro & 45.4 & 56.2 & 30.9 & 64.1 & 43.6 & 51.3 & 46.3 & 36.0 & 34.6 \\
    Gemini-2.5-Pro & 53.6 & 46.0 & 37.4 & 68.7 & 54.4 & 62.0 & 43.9 & \underline{47.4} & 68.8 \\
    \midrule
    \rowcolor{gray!10}
    \multicolumn{10}{l}{\textcolor{black}{\textit{Open-Sourced Models}}} \\
    LongVA-7B & 29.2 & 38.0 & 16.6 & 38.9 & 22.2 & 33.1 & 43.3 & 25.4 & 15.7 \\
    VILA-1.5-8B & 28.9 & 17.4 & 21.8 & 50.3 & 18.8 & 32.1 & 34.8 & 31.0 & 24.8 \\
    VILA-1.5-40B & 31.2 & 22.4 & 24.8 & 48.7 & 22.7 & 40.5 & 25.7 & 31.5 & 32.9 \\
    LLaVA-OneVision-7B & 32.4 & 47.7 & 20.2 & 47.4 & 12.3 & 42.5 & 35.2 & 29.4 & 24.4 \\
    LLaVA-OneVision-72B & 40.2 & 43.5 & 23.9 & 57.6 & 37.5 & 42.5 & 39.9 & 32.5 & 44.6 \\
    LLaVA-NeXT-Video-7B & 35.6 & 48.5 & 14.0 & 47.8 & 24.2 & 43.5 & 42.4 & 34.0 & 30.6 \\
    LLaVA-NeXT-Video-72B$\ $ & 40.9 & 48.9 & 22.8 & 57.4 & 35.3 & 42.4 & 36.7 & 35.0 & 48.6 \\
    Qwen2.5-VL-7B & 33.0 & 40.9 & 14.8 & 43.4 & 10.7 & 38.6 & 38.5 & 33.0 & 29.8 \\
    Qwen2.5-VL-72B & 37.0 & 25.1 & 29.3 & 54.5 & 38.8 & 38.2 & 37.0 & 34.0 & 28.9 \\
    InternVL3-8B & 42.1 & 68.1 & 39.0 & 48.4 & 33.6 & 48.3 & 36.4 & 27.3 & 35.4 \\
    InternVL3-78B & 48.4 & 71.2 & \underline{53.7} & 44.4 & 39.5 & 55.9 & 39.5 & 28.9 & 54.5 \\
    \midrule
    \rowcolor{gray!10}
    \multicolumn{10}{l}{\textcolor{black}{\textit{Spatial-Enhanced Models}}} \\
    Spatial-MLLM-4B & 48.4 & 65.3 & 34.8  & 63.1 & 45.1 & 41.3 & 46.2 & 33.5 & 46.3 \\
    VG-LLM-4B & 47.3 & 66.0 & 37.8  & 55.2 & 59.2 & 44.6 & 45.6 & 33.5 & 36.4 \\
    VG-LLM-8B & 50.7 & 67.9 & 37.7  & 58.6 & 62.0 & 46.6 & 40.7 & 32.4 & 59.2 \\
    Cambrian-S-3B & 57.3 & \underline{70.7} & 40.6 & 68.0 & 46.3 & 64.8 & 61.9 & 27.3 & \underline{78.8} \\
    VLM-3R-7B & \underline{60.9} & 70.2 & 49.4 & \underline{69.2} & \underline{67.1} & \underline{65.4} & \underline{80.5} & 45.4 & 40.1 \\
    \rowcolor{gray!10}
    \textbf{GeoAlign-4B~(Ours)} & \textbf{71.4} & \textbf{71.2} & \textbf{59.8} & \textbf{74.1} & \textbf{75.0} & \textbf{72.0} & \textbf{87.1} & \textbf{50.5} & \textbf{81.7} \\
    \bottomrule
    \end{tabular}
    }
    \label{tab:vsi_bench_results}
\end{table*}

\section{Experiments}
\label{sec:experiments}

To evaluate our proposed GeoAlign method, we conduct comprehensive experiments. In Sec.~\ref{sec:exp_implementation}, we provide specific implementation details. In Sec.~\ref{sec:exp_spatial_reasoning} and Sec.~\ref{sec:exp_3d_scene}, we assess GeoAlign on spatial reasoning and 3D scene understanding benchmarks. The results demonstrate that our method effectively mitigates task misalignment bias, achieving state-of-the-art performance among comparable models. In Sec.~\ref{sec:exp_ablations}, we provide extensive ablation studies to validate the specific configurations of each architectural component in our approach.

\subsection{Implementation}
\label{sec:exp_implementation}

\paragraph{Model.} 
Our GeoAlign is a compact model implemented upon widely used foundation models. For the multimodal large language model, we employ Qwen2.5-VL-3B~\cite{bai2025qwen2.5vl}. For the geometric encoder, we adopt the VGGT model~\cite{wang2025vggt}. To capture a comprehensive spectrum of geometric structures, we extract features from the latter half of VGGT layers~(12 layers in total). During the layer-wise routing phase, the sparsity hyperparameter is set to $K=2$.

\paragraph{Training.}
The model is trained for a single epoch on an empirically aggregated dataset comprising 460K samples. During the training process, the vision encoders~(both the Qwen2.5-ViT and the VGGT) are frozen to protect robust pretrained representations, while the feature fusion module and the language model are trainable. We utilize the AdamW optimizer, with a batch size of 64 and a uniform learning rate of 1e-5. To ensure stability, we apply a cosine learning rate decay schedule with a brief linear warmup phase covering the first 3\% training steps. All experiments are conducted on 8 NVIDIA H800 GPUs with DeepSpeed ZeRO Stage-2 optimization in BFloat16 precision.

\subsection{Spatial Reasoning}
\label{sec:exp_spatial_reasoning}

\paragraph{Datasets and Metrics.}
We assess the spatial reasoning capabilities on VSI-Bench~\cite{yang2025thinking}. VSI-Bench is sourced from ScanNet~\cite{dai2017scannet}, ScanNet++~\cite{yeshwanth2023scannet++}, and ARKitScenes~\cite{dehghan2021arkitscenes}, comprising over 5,000 QA samples across 8 different tasks. Following the benchmark guidelines, we measure the accuracy for multiple-choice tasks and the mean relative accuracy for numerical tasks.

\paragraph{Baselines.}
We compare GeoAlign against a wide range of representative models, including: proprietary models GPT-4o~\cite{hurst2024gpt4o}, Gemini-1.5-Pro~\cite{team2024gemini}, and Gemini-2.5-Pro~\cite{comanici2025gemini}; general-purpose open-source models LongVA~\cite{zhang2024longva}, VILA-1.5~\cite{lin2024vila}, LLaVA-OneVision~\cite{li2024llava}, LLaVA-NeXT-Video~\cite{zhang2024llavanext}, Qwen2.5-VL~\cite{bai2025qwen2.5vl}, and InternVL3~\cite{zhu2025internvl3}; recent spatial-enhanced models Spatial-MLLM~\cite{wu2025spatialmllm}, VG-LLM~\cite{zheng2025vgllm}, Cambrian-S~\cite{yang2025cambrian}, and VLM-3R~\cite{fan2026vlm3r}.

\paragraph{Results.}
Table~\ref{tab:vsi_bench_results} shows the quantitative results on VSI-Bench. Our GeoAlign model achieves state-of-the-art performance with a remarkable average score of 71.4. Crucially, despite operating at a compact 4B scale, GeoAlign demonstrates exceptional parameter efficiency and significantly eclipses larger proprietary models and open-source models, yielding a substantial improvement of over 10\% compared to the previous leading model.
Meanwhile, GeoAlign exhibits comprehensive capabilities across disparate spatial tasks from precise observation~(e.g., \emph{absolute distance} task reaching 59.8) to global understanding~(e.g., \emph{room size} task reaching 75.0). This balanced improvement empirically confirms that our proposed method effectively empowers the MLLM to break the performance bottleneck of static single-layer extraction.

\subsection{3D Scene Understanding}
\label{sec:exp_3d_scene}

\paragraph{Datasets and Metrics.}
To further assess the 3D scene understanding capabilities, we conduct evaluation on the 3D question-answering benchmarks ScanQA~\cite{azuma2022scanqa} and SQA3D~\cite{masqa3d}. Both datasets are built upon the ScanNet~\cite{dai2017scannet} scenes. We adhere to the standard evaluation protocols for each benchmark. For ScanQA, we measure the generation quality using four linguistic metrics: BLEU-4, METEOR, ROUGE-L, and CIDEr. For SQA3D, we measure the exact match accuracy~(EM-1).

\paragraph{Baselines.}
We select representative baseline models across three distinct categories: task-specific models specifically trained for 3D question-answering, including ScanQA~\cite{azuma2022scanqa}, SQA3D~\cite{masqa3d}, and 3D-VisTA~\cite{zhu20233dvista}; 3D/2.5D-input models that demand explicit geometric inputs~(e.g., point clouds or depth maps), including 3D-LLM~\cite{hong20233dllm}, LL3DA~\cite{chen2024ll3da}, ChatScene~\cite{zhang2024chatscene}, 3D-LLaVA~\cite{deng20253dllava}, Video-3D-LLM~\cite{zheng2025video3dllm}, and LLaVA-3D~\cite{zhu2025llava3d}; video-input models that do not require explicit 3D input, including Qwen2.5-VL~\cite{bai2025qwen2.5vl}, LLaVA-Video~\cite{zhang2024llavavideo}, Oryx-34B~\cite{liu2025oryx}, Spatial-MLLM~\cite{wu2025spatialmllm}, and VLM-3R~\cite{fan2026vlm3r}.

\paragraph{Results.}
The quantitative results on the ScanQA and SQA3D benchmarks are presented in Table~\ref{tab:scanqa_results}. Relying solely on video inputs without any explicit 3D data, our GeoAlign exhibits highly competitive capabilities. Compared to the leading video-input model VLM-3R, GeoAlign achieves closely comparable performance while utilizing a compact size of barely half the parameters. This demonstrates the efficacy of our proposed GeoAlign mechanism in extracting crucial geometric features for 3D scene understanding, achieving high parameter efficiency without complex modules.

\begin{table}[!t]
\centering
\caption{\textbf{Evaluation on ScanQA and SQA3D for 3D scene understanding.} In this table, ``B-4'', ``M'', ``R'', ``C'', and ``EM-1'' denote BLEU-4, METEOR, ROUGE-L, CIDEr, and exact match accuracy, respectively. Among video-input models, the best performance is in \textbf{bold}, and the second best is \underline{underlined}.}
\setlength\tabcolsep{4pt} 
\resizebox{\columnwidth}{!} 
{
    \begin{tabular}{l|cccc|c}
    \toprule
    \multirow{2}{*}{\raisebox{-0.5ex}{\textbf{Models}}}
    & \multicolumn{4}{c|}{\textbf{ScanQA}}
    & \textbf{$\ \ $SQA3D$\ \ $}
    \\
    \cmidrule(lr){2-5} \cmidrule(lr){6-6}
    & $\ \ $B-4$\ \ $ & M & R & C & EM-1 \\
    \midrule 
    \rowcolor{gray!10}
    \multicolumn{6}{l}{\textcolor{black}{\textit{Task-Specific Models}}} \\
    ScanQA & 10.1 & 13.1 & 33.3 & 64.9 & 47.2  \\
    SQA3D & 11.2 & 13.5 & 34.5 & - & 46.6  \\
    3D-VisTA & 10.4 & 13.9 & 35.7 & 69.6 & 48.5 \\
    \midrule
    \rowcolor{gray!10}
    \multicolumn{6}{l}{\textcolor{black}{\textit{3D/2.5D-Input Models}}} \\
    3D-LLM & 12.0 & 14.5 & 35.7 & 69.4 & -  \\
    LL3DA & 13.5 & 15.9 & 37.3 & 76.8 & -  \\
    ChatScene & 14.3 & 18.0 & 41.6 & 87.7 & 54.6 \\
    3D-LLaVA & 17.1 & 18.4 & 43.1 & 92.6 & 54.5 \\
    Video-3D-LLM & 16.4 & 20.0 & 49.3 & 102.1 & 58.6\\
    LLaVA-3D & 16.4 & 20.8 & 49.6 & 103.1 & 60.1 \\
    \midrule 
    \rowcolor{gray!10}
    \multicolumn{6}{l}{\textcolor{black}{\textit{Video-Input Models}}} \\
    Qwen2.5-VL-7B & 8.0 & 11.4 & 29.3 & 53.9 & 46.5 \\
    Qwen2.5-VL-72B & 12.0 & 13.0 & 35.2 & 66.9 & 47.0 \\
    LLaVA-Video-7B & 3.1 & 17.7 & 44.6 & 88.7 & 48.5\\
    Oryx-34B & - & 15.0 & 37.3 & 72.3 & 50.9 \\
    Spatial-MLLM-4B & 14.8 & 18.4 & 45.0 & 91.8 & 55.9 \\
    VLM-3R-7B & \underline{15.5} & \textbf{19.7} & \textbf{49.1} & \textbf{101.9} & \textbf{60.7} \\
    \rowcolor{gray!15}
    \textbf{GeoAlign-4B~(Ours)$\ $}  & \textbf{15.7} & \underline{19.4} & \underline{48.2} & \underline{99.4} & \underline{60.3} \\
    \bottomrule
    \end{tabular}
}
\label{tab:scanqa_results}
\end{table}

\subsection{Ablation Studies}
\label{sec:exp_ablations}

To validate the specific design of our architecture, we conduct detailed ablation studies on the VSI-Bench. We use the same base models and training settings for all variants.


\begin{figure}[t]
  \centering
   \includegraphics[width=\columnwidth]{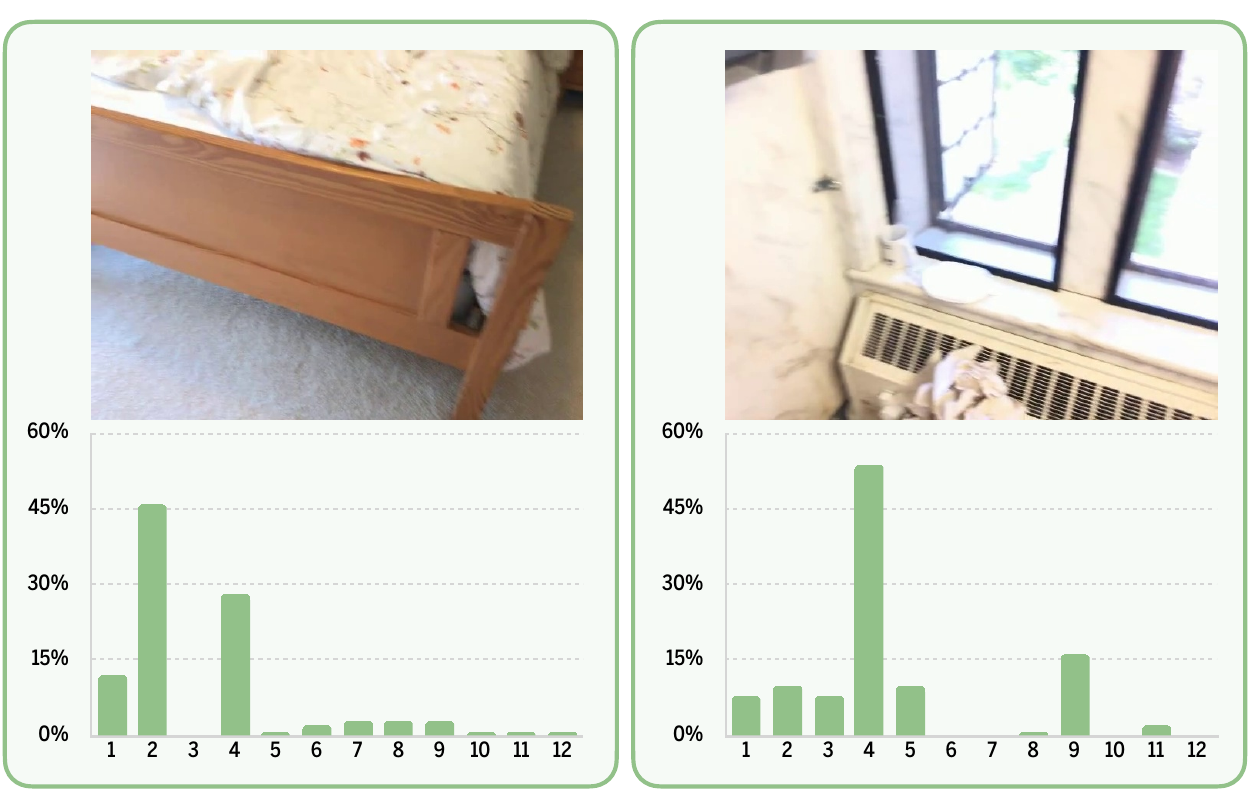}
   \caption{\textbf{Qualitative visualization of the dynamic routing mechanism.} We visualize the mean routing weights~($\bm{\alpha}$, presented as percentages) assigned to each layer of the geometric feature bank across the entire visual sequence. For distinct visual inputs, the routing distributions exhibit significant variations.}
   \label{fig:sample}
\end{figure}

\paragraph{Geometric Feature Usage.}
We ablate our dynamic routing mechanism against three distinct baseline strategies. The ``2D-Only'' baseline denotes directly fine-tuning the Qwen2.5-VL-3B model, using LoRA in the vision encoder while not injecting any geometric features. The ``Single'' setting statically injects deep geometric features from Layer-22 of VGGT. The ``Mean'' setting uniformly pools the geometric features across all 12 candidate layers prior to injection.
As shown in Table~\ref{tab:ablation}, integrating geometric features significantly enhances spatial reasoning capabilities, yet relying on a static single layer suffers from the task misalignment bias and leaves a performance gap. While mean pooling fusion improves the performance, it homogenizes diverse layers and fails to fully exploit the geometric features. In contrast, our proposed method resolves this dilemma, effectively reconciling multiple layers of geometric features to yield better overall performance.
As visualized in Fig.~\ref{fig:sample}, the dynamic routing distributions exhibit context-aware variations across distinct input scenes, confirming that GeoAlign achieves dynamic feature utilization.

\begin{table*}[!t]
\centering
\caption{\textbf{Ablation studies.} We evaluate key architectural settings: (1) Geometric Feature Usage, comparing our dynamic routing against baselines using no geometric features, single-layer geometric features, or mean geometric features; (2) Geometric Feature Selection, assessing the construction of geometric feature bank using different VGGT layers~(uniformly sampled, former half, latter half); (3) Injection Position, investigating injection stages from before the LLM to various internal layers; (4) Sparsity Hyperparameter, determining the number of selected layers~($K$) in our dynamic routing; (5) Projection and Fusion Mechanism, comparing our shared projector and direct residual addition against split, modulated, or gated variants. The best is in \textbf{bold}, and the second best is \underline{underlined}.}
\resizebox{\textwidth}{!}
{
    \begin{tabular}{l|c|cccccccc}
    \toprule
    Setting & $\ $Avg.$\ $ & Obj. Cnt. & Abs. Dist. & Obj. Size & Room Size & Rel. Dist. & Rel. Dir. & Route Plan & Appr. Order 
    \\
    \midrule
    \rowcolor{gray!10}
    \multicolumn{10}{l}{\textit{Geometric Feature Usage}} \\
    2D-Only & 66.8 & 70.3 & 49.6 & 73.5 & 68.3 & 64.2 & 78.1 & \underline{47.4} & \textbf{82.7} \\
    Single    & 69.3 & 71.0 & 58.2 & \underline{73.7} & 72.7 & 69.4 & \textbf{87.2} & 40.7 & 81.7 \\
    Mean & \underline{70.5} & \textbf{71.5} & \underline{58.3} & \underline{73.7} & \underline{73.4} & \underline{70.7} & \underline{87.1} & \underline{47.4} & \underline{81.9} \\
    \rowcolor{gray!10}
    \textbf{Dynamic} & \textbf{71.4} & \underline{71.2} & \textbf{59.8} & \textbf{74.1} & \textbf{75.0} & \textbf{72.0} & \underline{87.1} & \textbf{50.5} & 81.7 \\
    \midrule
    \rowcolor{gray!10}
    \multicolumn{10}{l}{\textit{Geometric Feature Selection}} \\
    Uniform & \underline{70.8} & \underline{70.7} & \underline{59.4} & \textbf{74.7} & \underline{73.8} & \underline{69.4} & \textbf{88.3} & 46.9 & \textbf{83.0} \\
    Former  & 67.2 & 70.6 & 50.0 & \underline{74.2} & 70.2 & 65.8 & 77.7 & \underline{47.4} & \underline{82.0} \\ 
    \rowcolor{gray!10}
    \textbf{Latter} & \textbf{71.4} & \textbf{71.2} & \textbf{59.8} & 74.1 & \textbf{75.0} & \textbf{72.0} & \underline{87.1} & \textbf{50.5} & 81.7 \\
    \midrule
    \rowcolor{gray!10}
    \multicolumn{10}{l}{\textit{Injection Position}} \\
    Early-LLM & 70.3 & 70.8 & 58.7 & 74.1 & 72.8 & 69.2 & \textbf{87.7} & 45.9 & \textbf{82.8} \\
    Mid-LLM   & 70.7 & \textbf{71.3} & 59.7 & \textbf{74.6} & \textbf{75.0} & 70.0 & 86.1 & \underline{50.0} & 79.3 \\
    Late-LLM  & 66.1 & 70.4 & 50.1 & 73.9 & 71.8 & 64.2 & 75.3 & 43.3 & 79.9 \\
    Multi-LLM & \underline{70.9} & 70.9 & \textbf{60.9} & \underline{74.2} & 74.0 & \textbf{72.1} & 86.9 & 47.9 & 80.7 \\
    \rowcolor{gray!10}
    \textbf{Pre-LLM} & \textbf{71.4} & \underline{71.2} & \underline{59.8} & 74.1 & \textbf{75.0} & \underline{72.0} & \underline{87.1} & \textbf{50.5} & \underline{81.7} \\
    \midrule
    \rowcolor{gray!10}
    \multicolumn{10}{l}{\textit{Sparsity Hyperparameter}} \\
    Top-1 & \underline{70.7} & \underline{70.5} & 57.0 & \textbf{74.2} & \underline{74.0} & \underline{70.8} & \textbf{89.6} & \underline{47.4} & \underline{81.7} \\ 
    Top-3 & 70.3 & 70.1 & \underline{59.3} & 73.9 & 73.3 & 69.0 & \underline{87.2} & 46.4 & \textbf{82.8} \\ 
    \rowcolor{gray!15}
    \textbf{Top-2} & \textbf{71.4} & \textbf{71.2} & \textbf{59.8} & \underline{74.1} & \textbf{75.0} & \textbf{72.0} & 87.1 & \textbf{50.5} & \underline{81.7} \\
    \midrule
    \rowcolor{gray!10}
    \multicolumn{10}{l}{\textit{Projection and Fusion Mechanism}} \\
    Split-Proj                        & 69.4 & 69.9 & 57.8 & \textbf{74.6} & 73.1 & 69.6 & 85.7 & 44.3 & 79.8 \\
    FiLM                        & 70.1 & \underline{70.5} & 57.9 & \textbf{74.6} & \textbf{75.6} & 69.3 & 84.7 & 45.9 & \textbf{82.5} \\
    Gated~(2D)       & \underline{70.7} & 70.1 & \textbf{59.8} & 74.1 & 73.6 & \underline{70.3} & \textbf{87.9} & \underline{47.4} & \textbf{82.5} \\
    Gated~(2D+3D)    & 70.2 & \underline{70.5} & 59.6 & 74.0 & 74.2 & 66.9 & \underline{87.7} & \underline{47.4} & 81.2 \\
    \rowcolor{gray!10}
    \textbf{Shared+Add}    & \textbf{71.4} & \textbf{71.2} & \textbf{59.8} & 74.1 & \underline{75.0} & \textbf{72.0} & 87.1 & \textbf{50.5} & 81.7 \\
    \bottomrule
    \end{tabular}
}
\label{tab:ablation}
\end{table*}

\paragraph{Geometric Feature Selection.}
We ablate the layer selection strategy for constructing the geometric feature bank. As shown in Table~\ref{tab:ablation}, we compare three configurations: uniformly sampled 12 layers, the former 12 layers, and the latter 12 layers. Among these, the latter 12 layers achieve the best performance. This comparison suggests that early stages still contain premature noise, while the latter half provides a more effective candidate pool for spatial reasoning tasks.

\paragraph{Injection Position.}
We investigate the geometric feature injection at various positions, including the early~(Layer-9), middle~(Layer-18), and late~(Layer-27) stages of LLM, a multi-layer combination~(Layer-9, 18, 27), as well as before LLM. For injections within the LLM backbone, we utilize the visual tokens from the corresponding layer as routing queries to ensure proper contextual alignment. The results in Table~\ref{tab:ablation} reveal a performance deterioration when the injection position moves into LLM. While distributing the injection across multiple layers recovers the performance to 70.9, it introduces more computational overhead without surpassing the pre-LLM injection. This empirically demonstrates that geometric features are better suited for enriching visual features rather than being injected into the stage of abstract semantics.

\paragraph{Sparsity Hyperparameter.}
A critical component of our dynamic routing mechanism is the strict Top-$K$ sparsity constraint, which insulates the MLLM from the interference of redundant geometric features. To determine the optimal sparsity, we ablate the value of $K$ in Table~\ref{tab:ablation}.
Setting $K=1$ enforces absolute purity but restricts the representational capacity, yielding declined performance. Conversely, relaxing the sparsity to $K=3$ also leads to a performance drop. Our selected $K=2$ strikes a good balance.

\paragraph{Projection and Fusion Mechanism.}
First, we compare our shared MLP projector $f_{\phi}(\cdot)$ in constructing the geometric feature bank against a split approach where each layer corresponds to an independent MLP projector. As shown in Table~\ref{tab:ablation}, the shared design yields significantly better performance than the split variant.
Second, we compare our straightforward residual addition ($\hat{\bm{Q}} = \bm{Q} + \bm{W}_{out}\hat{\bm{F}}$) against three variants:
(1) FiLM, a feature-wise linear modulation~\cite{perez2018film} where the original visual token $\bm{Q}$ is scaled and shifted by $\hat{\bm{F}}$, formulated as $\hat{\bm{Q}} = \bm{Q} \odot (\mathbf{1} + \bm{W}_{scale}\hat{\bm{F}}) + \bm{W}_{shift}\hat{\bm{F}}$; (2) Gated~(2D), a patch-level gate predicted by the 2D visual tokens, given by $\hat{\bm{Q}} = \bm{Q} + \bm{W}_{out}\big(\sigma(\bm{W}_{gate}\bm{Q}) \odot \hat{\bm{F}}\big)$; and (3) Gated~(2D+3D), a patch-level gate predicted by concatenating 2D and 3D features, computed as $\hat{\bm{Q}} = \bm{Q} + \bm{W}_{out}\big(\sigma(\bm{W}_{gate}[\bm{Q}; \hat{\bm{F}}]) \odot \hat{\bm{F}}\big)$.
These two ablations suggest avoiding parameter redundancy and complex architectures, thereby enabling the model to stably and spontaneously learn to utilize geometric features.


\section{Conclusion}
\label{sec:conclusion}
In this paper, we present GeoAlign, a novel framework that empowers MLLMs with robust spatial reasoning capabilities. We first identify a critical task misalignment bias prevalent in existing spatial-enhanced MLLMs, wherein the static extraction of a single deep geometric layer fundamentally contradicts diverse spatial demands. To overcome this, GeoAlign introduces feature realignment for spatial reasoning. By constructing a hierarchical geometric feature bank and leveraging the MLLM's original visual tokens as active queries, our method performs dynamic layer-wise sparse routing to adaptively fetch the suitable geometric features for each patch. Extensive experiments across VSI-Bench, ScanQA, and SQA3D demonstrate the superiority of our approach, and empirically validate the specific architectural configurations.

\section*{Limitations}
\label{sec:limitations}
\paragraph{Geometric Foundation Model.} While GeoAlign effectively mitigates the task misalignment bias, it relies on a frozen, off-the-shelf 3D foundation model to provide geometric features. However, this model is not natively tailored and trained for MLLMs' spatial reasoning demands. Consequently, the extracted geometric features may still face both information insufficiency for complex spatial tasks and task-irrelevant redundancies. Furthermore, maintaining a large-scale 3D foundation model exclusively for feature extraction incurs additional parameter overhead. 

\paragraph{Computational Overhead.} The dynamic layer-wise routing mechanism, while effective in realigning with the task demands, inevitably incurs additional computational overhead during the forward pass. To construct the comprehensive candidate pool, multiple intermediate layers must be extracted and maintained in GPU memory. Compared to static single-layer extraction, this inherently increases the memory footprint.

\bibliography{custom}




\end{document}